\documentclass{article}
\usepackage{amsmath,graphicx,mlspconf}
\usepackage{amssymb}
\usepackage{float}
\usepackage{subcaption}
\usepackage{multirow}
\usepackage{pgfplots, soul} 

\newcommand{\imagepath}{./images}

\newcommand{\code}{\mathbf{x}_t}
\newcommand{\prev}{\mathbf{x}_{t-1}}
\newcommand{\entropy}{\mathrm{H}}
\newcommand{\refframe}{\hat{\mathbf{x}}_{t-1}}
\newcommand{\balph}{\boldsymbol{\alpha}}
\newcommand{\sysoutput}{\hat{\mathbf{x}}_t}
\newcommand{\prediction}{\tilde{\mathbf{x}}_t}
\newcommand{\diagramscale}{0.265}
\newcommand{\staticarea}{\mathcal{S}}
\newcommand{\nonstaticarea}{\bar{\staticarea}}

\copyrightnotice{978-1-7281-6662-9/20/\$31.00 {\copyright}2020 IEEE}

\toappear{2020 IEEE International Workshop on Machine Learning for Signal Processing, Sept.\ 21--24, 2020, Espoo, Finland}

\title{ModeNet: MODE SELECTION NETWORK FOR LEARNED VIDEO CODING}
\name{Th\'{e}o LADUNE\textsuperscript{*$\dagger$}, Pierrick PHILIPPE\textsuperscript{*}, Wassim HAMIDOUCHE\textsuperscript{$\dagger$}, Lu ZHANG\textsuperscript{$\dagger$}, Olivier D\'{E}FORGES\textsuperscript{$\dagger$}}
\address{\textsuperscript{*}Orange Labs, France\\
\textsuperscript{$\dagger$}Univ. Rennes, INSA Rennes,
CNRS, IETR --  UMR 6164, Rennes, France\\
\normalsize{\texttt{theo.ladune@orange.com}}\\
}

\begin{document}

\maketitle

\begin{abstract}
In this paper, a mode selection network (ModeNet) is proposed to enhance
deep learning-based video compression. Inspired by traditional video coding, ModeNet
purpose is to enable competition among several coding modes.

The proposed ModeNet learns and conveys a pixel-wise partitioning of the frame,
used to assign each pixel to the most suited coding mode. ModeNet is trained
alongside the different coding modes to minimize a rate-distortion cost. It is a
flexible component which can be generalized to other systems to allow competition
between different coding tools. ModeNet interest is studied on a P-frame coding
task, where it is used to design a method for coding a frame given its
prediction. ModeNet-based systems achieve compelling performance when evaluated
under the \textit{Challenge on Learned Image Compression 2020} (CLIC20) P-frame
coding track conditions.
\end{abstract}
\section{Introduction and Related Works}
\label{sec:intro}

Modern video compression systems widely adopt coding mode competition to select
the best performing tool given the signal. Coding performance improvements
of MPEG/ITU video codecs (AVC, HEVC and VVC) \cite{DBLP:journals/cm/MarpeWS06,
Sullivan:2012:OHE:2709080.2709221, VVC_Ref} are mainly brought by increasing the number
of coding modes. These modes include prediction mode (Intra/Inter), transform
type and block shape. This concept allows to perform signal adapted processing. 

In recent years, image coding standards such as BPG (HEVC-based image coding
method) have been outperformed by neural networks-based systems
\cite{DBLP:conf/nips/MinnenBT18,
DBLP:conf/iclr/LeeCB19,DBLP:journals/corr/abs-2002-03370}. Most neural
networks-based systems are inspired by Ball\'{e} and Minnen's
works~\cite{DBLP:conf/nips/MinnenBT18, DBLP:conf/iclr/BalleLS17,
DBLP:conf/iclr/BalleMSHJ18}. They rely on an Auto-Encoder (AE)
architecture that maps the input signal to latent variables. Latent variables
are then transmitted with entropy coding, based on a probability model conveyed
as an Hyper-Prior (HP). Such systems are denoted as AE-HP systems in the
remaining of the paper. AE-HP systems are learned in an end-to-end fashion: all
components being trained according to a global objective function, minimizing a
trade off between distortion and rate. Training of AE-HP systems is often
performed following Ball\'{e}'s method \cite{DBLP:conf/iclr/BalleLS17} to
circumvent the presence of non-differentiable elements in the auto-encoder.

As learned image compression already exhibits state-of-the-art performance,
learned video compression has started to attract the research community's
attention. Authors in
\cite{DBLP:conf/cvpr/LuO0ZCG19,DBLP:conf/iccv/DjelouahCSS19} proposed a method
to compress Groups Of Pictures (GOPs) inspired by standard video coding methods
\textit{i.e.} by decomposing GOPs into intra frames, without dependency, and
inter frames which are coded based on previously decoded frames. Intra frames
are coded with AE-HP systems while Inter frames processing is widely inspired by
classical codecs approaches, replacing traditional coding tools by neural
networks. First, motion vectors (representing the motion between the current
frame and the reference frames), are estimated by an optical flow
network~\cite{DBLP:conf/cvpr/RanjanB17,DBLP:conf/cvpr/SunY0K18}. Motion vectors
are encoded using a AE-HP system and used to perform a prediction of the current
frame. Finally, the residue (prediction error) is computed either in image or
latent domain and coded using an other AE-HP system. Liu \textit{et al.}
\cite{DBLP:journals/corr/abs-1912-06348} tackle a similar problem and show that
using a single network for both flow estimation and coding achieves performance
similar to HEVC.

Although learned video coding already demonstrates appealing performance, it
does not exploit all usual video coding tools. Particularly, inter frames are fully
transmitted through motion compensation and residual coding even though it may
not be the best option. This is different from classical encoders,
where inter frame coding relies on a combination of \textit{Skip Mode} (direct
copy of the motion compensated reference), intra coding and residual inter
coding.

In this paper, a mode selection network (ModeNet) is proposed. Its role is to
select the most suited coding mode for each pixel. ModeNet is based on a
lightweight AE-HP system, which is trained end-to-end alongside the networks
performing the different coding modes. It learns to assign each pixel to the
coding mode that provides the best rate-distortion tradeoff. Consequently, the
proposed ModeNet can be integrated seamlessly into any neural-based coding
scheme to select the most appropriate coding mode.

ModeNet behavior and benefits are highlighted through an architecture
composed of two competing modes: each pixel is either copied from the prediction
(\textit{Skip Mode} in classical codecs) or conveyed through an AE-HP coder. We show
that using ModeNet achieves compelling performance when evaluated under the
\textit{Challenge on Learned Image Compression 2020} (CLIC20) P-frame coding
track conditions \cite{CLIC20_web_page}. 

\begin{figure}
    \centering

    \includegraphics[scale=0.40]{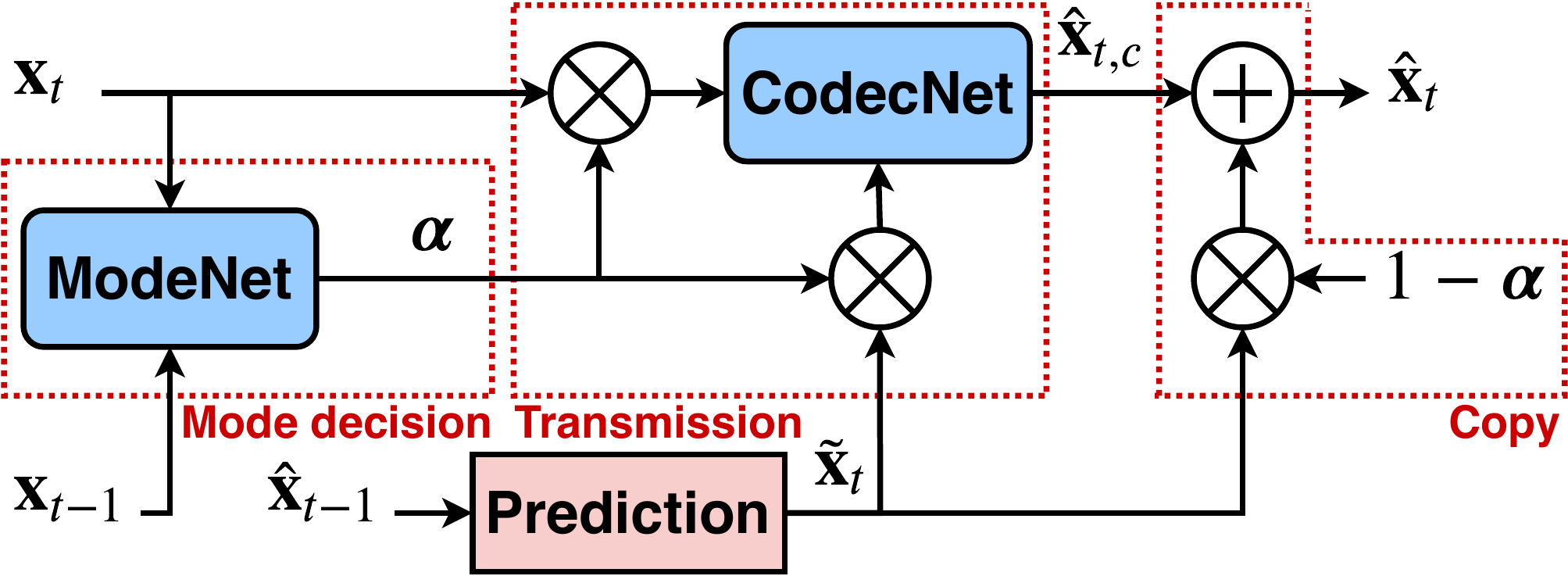}
    \caption{Architecture of the complete system.}
    \label{CompleteSystemDiagrams}
\end{figure}

\section{Preliminary}

This work focuses on \textit{P-frame coding} with two \textit{AE-HP systems}.
Theses two concepts are briefly summarized below.

\textbf{AE-HP system} This coding scheme is composed of a convolutional encoder
which maps the input signal to latents and a convolutional decoder which
reconstructs the input signal from quantized latents. Latents are transmitted
with entropy coding based on a latents probability model. To improve
performance, the probability model is conditioned on
side-information~\cite{DBLP:conf/iclr/BalleMSHJ18} and/or on previously received
latents~\cite{DBLP:conf/nips/MinnenBT18}.

\textbf{P-frame coding} Let $\left(\prev,\code\right) \in \mathbb{R}^{2 \times C
\times H \times W}$ be the previous frame and the frame to be coded,
respectively. $C$, $H$ and $W$ denote the number of color channels, height and
width of the image, respectively. The previous frame $\prev$ has already been
transmitted and it thus available at the decoder side to be used as a reference
frame $\refframe$. Since this work follows the CLIC20 P-frame coding test
conditions, the coding of $\prev$ frame is considered lossless \textit{i.e.}
$\refframe = \prev$. 
P-frame coding is the process to encode $\code$ knowing $\refframe$. A
prediction $\prediction$ of $\code$ is made available, based on $\refframe$ and
side-information (such as motion). The conditional entropy of $\code$ and $\prediction$ verifies:
\begin{equation}
    \entropy(\code \mid \prediction) = \entropy(\code) - \mathrm{I}(\code, \prediction) \leq \entropy(\code),
\end{equation}
where $\entropy$ is the Shannon entropy and $\mathrm{I}$ is the mutual
information. Thus using information from $\prediction$ allows to lower the
uncertainty about $\code$, resulting in better coding performance. This work
aims at minimizing a rate-distortion trade-off under a lossy P-frame coding objective:
\begin{equation}
    \mathcal{L}(\lambda) = \mathrm{D}(\sysoutput, \code) + \lambda \, R,\ \text{with}\ \sysoutput = f(\prediction, \code),
\end{equation}
where $\mathrm{D}$ is a distortion measure, $\sysoutput$ is the reconstruction
from an encoding/decoding process $f$ with an associated rate $R$ weighted by a
Lagrange multiplier $\lambda$.
\begin{figure*}[htb]
    \begin{subfigure}[b]{1\columnwidth}
        \centering
        \includegraphics[scale=\diagramscale]{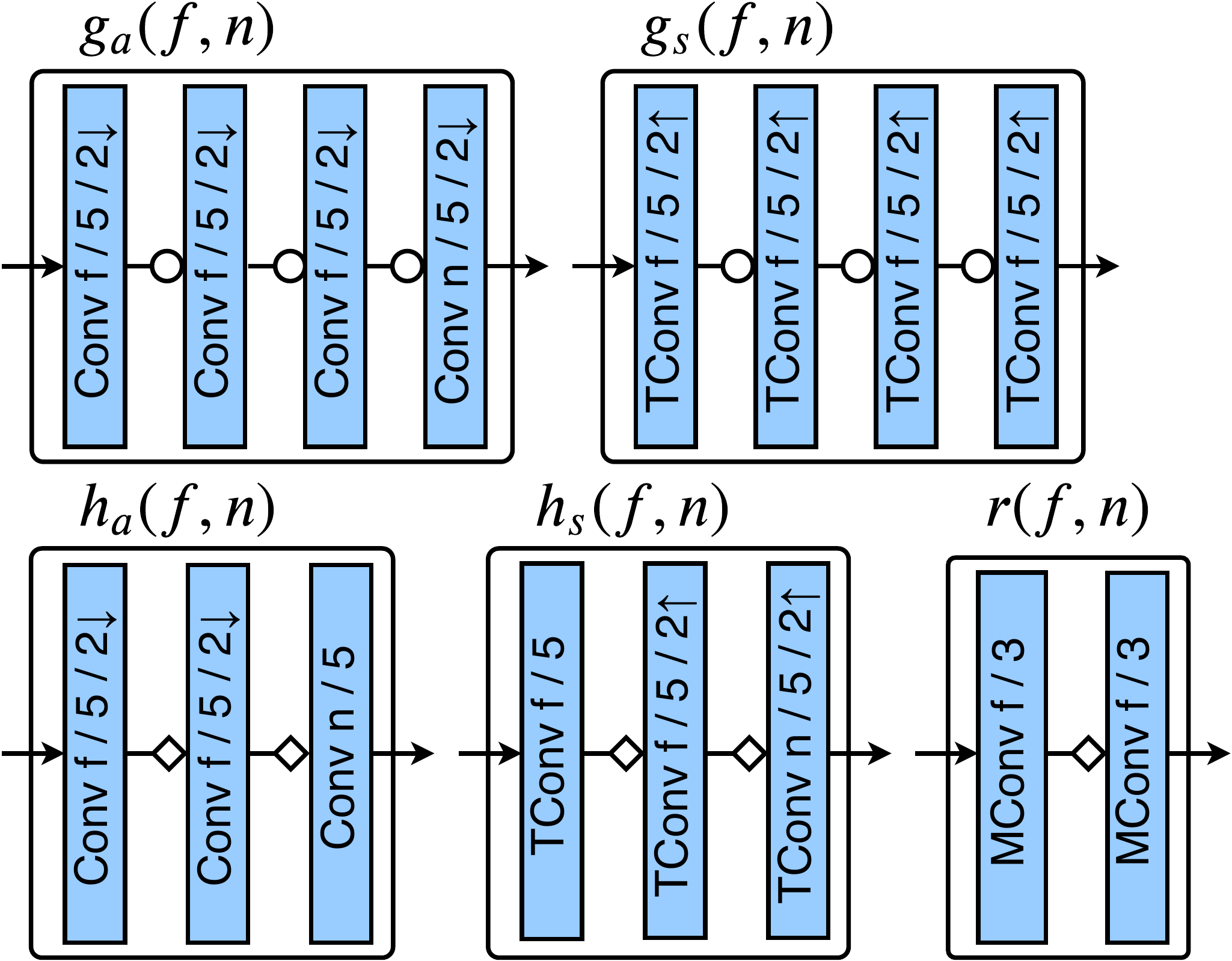}
        \caption{Basic blocks used to build AE-HP systems.}
        \label{RawBlocksDiagrams}
    \end{subfigure}
    \hfill
    \begin{subfigure}[b]{1\columnwidth}
        \centering
        \includegraphics[scale=\diagramscale]{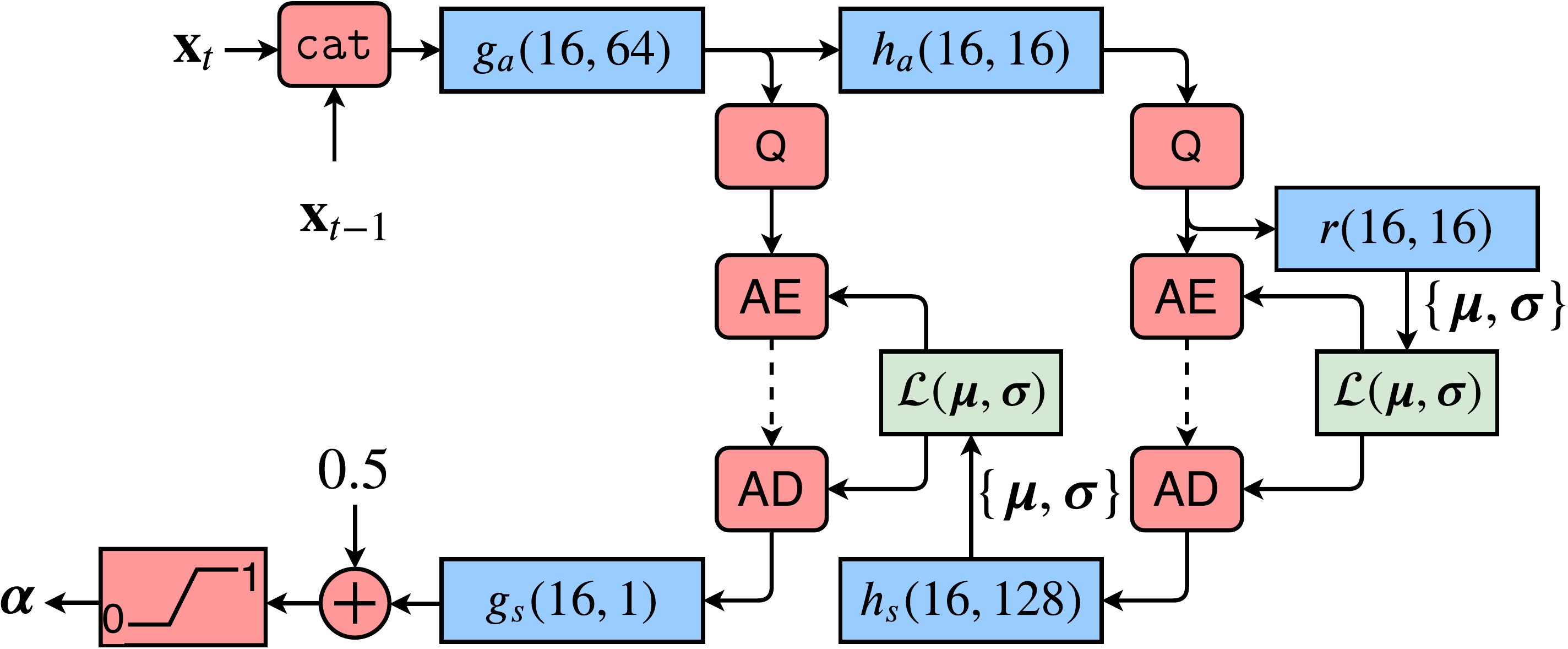}
        \caption{ModeNet architecture. $g_a$ and $g_s$ use LeakyReLU.}
        \label{ModeNetDiagram}
    \end{subfigure}

    \begin{subfigure}[b]{1\columnwidth}
        \centering
        \includegraphics[scale=\diagramscale]{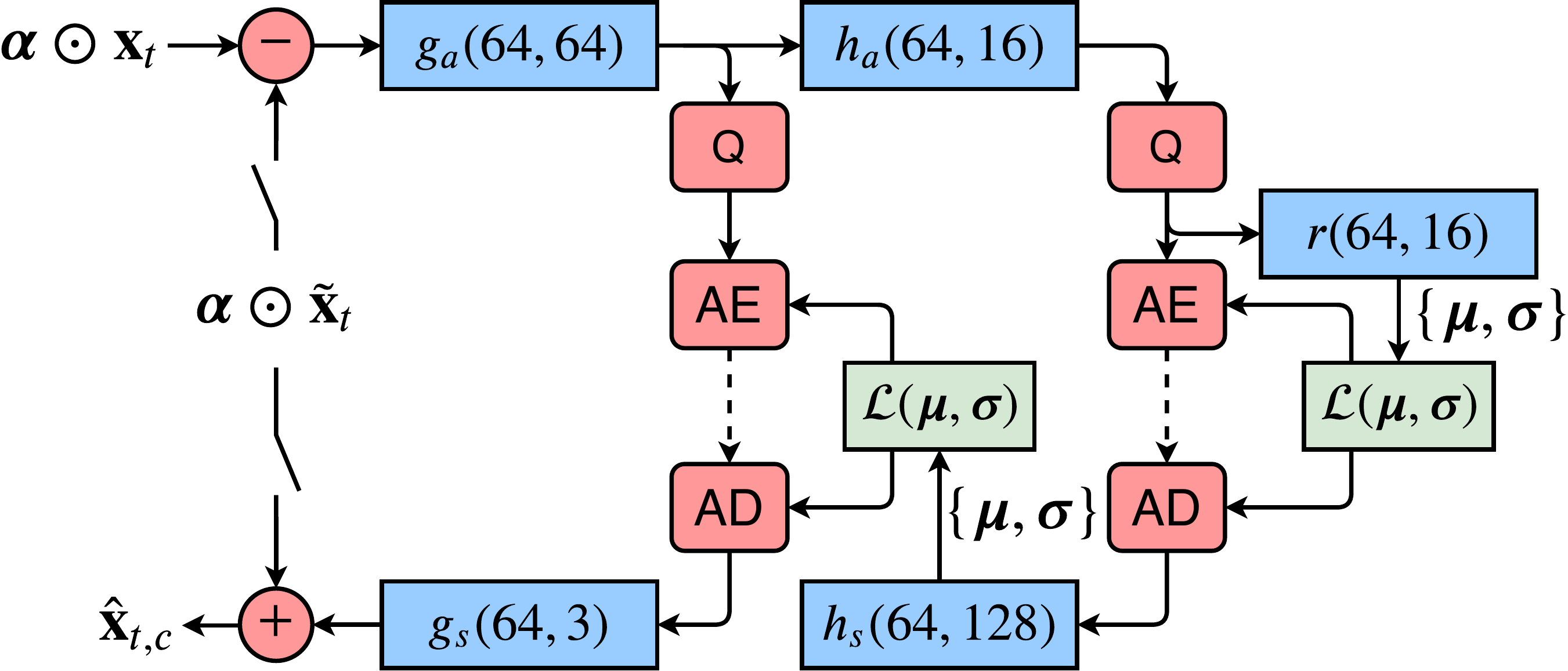}
        \caption{CodecNet for image and difference coding. $g_a$ and $g_s$ use GDN.}
        \label{CodecNetImageDiffDiagram}
    \end{subfigure}
    \hfill
    \begin{subfigure}[b]{1\columnwidth}
        \centering
        \includegraphics[scale=\diagramscale]{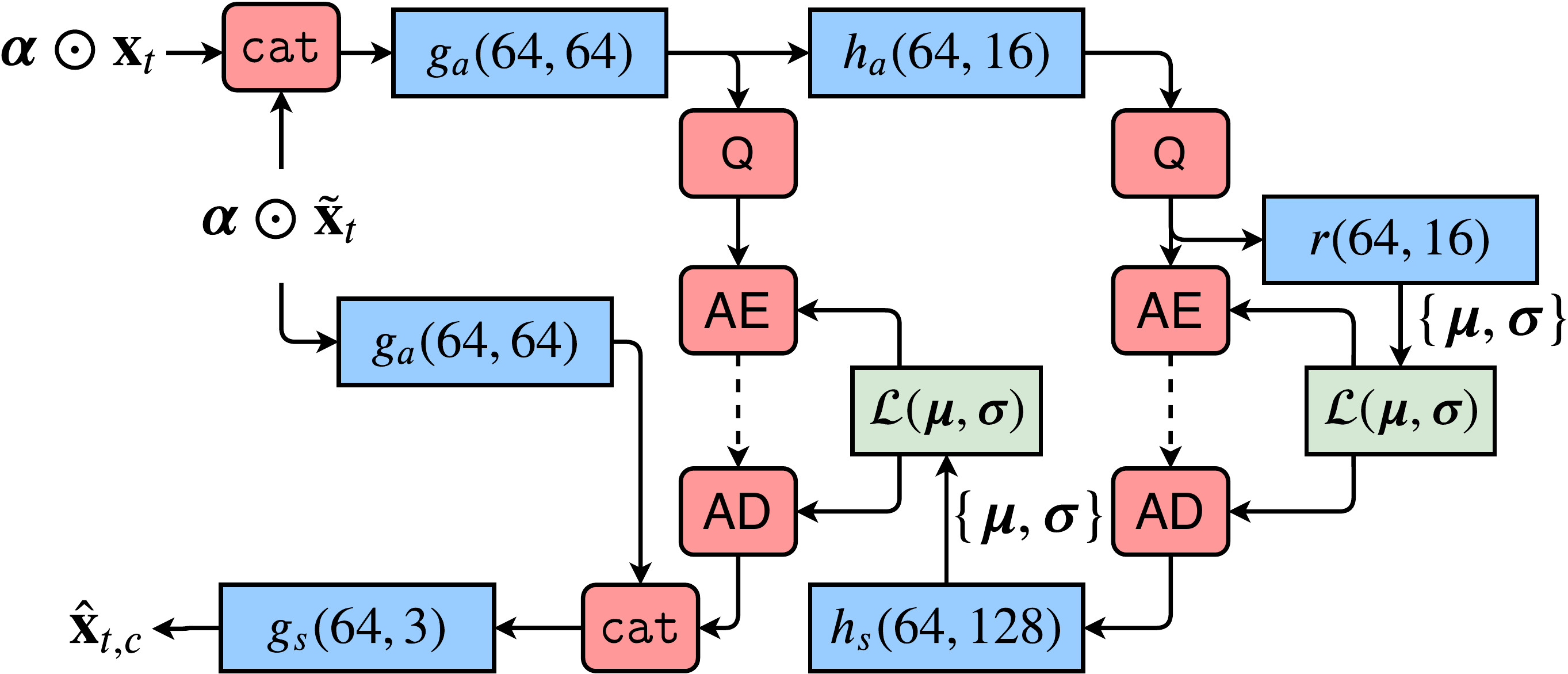}
        \caption{CodecNet for conditional coding. $g_a$ and $g_s$ use GDN.}
        \label{CodecNetConditionalDiagram}
    \end{subfigure}
    \caption{Detailed architecture of all used networks. \textbf{Top left figure:}
    Building blocks of all subsystems. $g_a$ and $g_s$ are the main
    encoder/decoder. $h_a$ and $h_s$ are the hyperprior encoder/decoder. $r$ is
    an auto-regressive module as in \cite{DBLP:conf/nips/MinnenBT18}. Each block
    is set-up by $f$ (number of internal features) and $n$ (number of output
    features). Squared arrows denote LeakyReLU, rounded arrows refer to either
    LeakyReLU or GDN \cite{DBLP:conf/iclr/BalleLS17}. Convolutions parameters:
    filters number $\times$ kernel size / stride. TConv and MConv stand
    respectively for Transposed convolution and Masked convolution.
    $\texttt{cat}$ stands for concatenation along feature axis, Q for
    quantization, AE and AD for arithmetic encoding/decoding with a Laplace
    distribution $\mathcal{L}$.}
    \label{fig:four figures}
\end{figure*}

\section{Mode Selection for P-frame Coding}
\label{sec:system}

\subsection{Problem formulation}

Let us define $\staticarea$ as a set of pixels of frame $\code$ verifying the following inequality:

\begin{equation}
    d(\prediction, \code;i) \leq d(\sysoutput, \code;i) + \lambda \, r(\code \mid \prediction;i),
    \label{eq:pixelWiseLossComparison}
\end{equation}
where ${d(\cdot,\cdot;i)}$ is the $i$-th pixel distortion and
${r(\code\mid\prediction;i)}$ the rate of the $i$-th pixel of $\code$ knowing
$\prediction$. The set $\staticarea$ gives the zones of $\code$ preferably
conveyed by using $\prediction$ copy (\textit{Skip Mode}) rather than by an
encoder-decoder system. $\staticarea$ is re-written as:
\begin{align}
    \begin{split}
    \staticarea = \left\{x_{t, i} \mid\ x_{t, i} \in \code,\ \ell (\prediction, \code; i) \leq \lambda \right\},\\
    \text{ with } \ell(\prediction, \code; i) = \frac{d(\prediction, \code;i) - d(\sysoutput, \code;i)}{r(\code \mid \prediction;i)}.
    \label{eq:staticArea}
    \end{split}
\end{align}
The partitioning function $\ell$ is a rate-distortion comparator, which assigns
a coding mode (either copy or transmission) to each pixel. It is
similar to the RD-cost used to arbitrate different coding modes in traditional
video coding. In the remaining of the article, $\nonstaticarea$ is the complement set of
$\staticarea$, used to denote all pixels not in $\staticarea$ \textit{i.e.}
pixels for which transmission results in a better rate-distortion trade-off than
copy from $\prediction$.

Hand-crafting the partitioning function $\ell$ is not trivial. Indeed, both the
rate and the distortion of the $i$-th pixel depends on choices made for previous
and future pixels. Classical codecs circumvent this issue by computing
$\ell$ on blocks of pixels assumed independent from each others.

The purpose of this work is to introduce a convolutional mode selection network
(ModeNet), whose role is both to indicate which pixels belong to $\staticarea$
and to convey this partitioning. This performs a pixel-wise
partitioning of $\code$, allowing both causal and anti-causal dependencies, learned
by minimizing a global rate-distortion objective function.

\subsection{Proposed system}

The proposed system is built around ModeNet, which learns a pixel-wise weighting
$\balph$ allowing to choose among two different coding methods for each pixel.
Here, the two methods in competition are copying prediction pixel from
$\prediction$ or coding pixels of $\code$ using an AE-HP system (CodecNet).

An overview of the system architecture is shown in Fig.
\ref{CompleteSystemDiagrams}. ModeNet and CodecNet architecture are described in
details in section \ref{subsec:detailArchitecture}. ModeNet is defined as a
function $m$:
\begin{equation}
    R_{m},\ \balph = m\left(\prev,\code\right),
\end{equation}
where $\balph \in \left[0 ; 1\right]^{H \times W}$ is the pixel-wise weighting
and $R_m$ the rate needed to convey $\balph$. The pixel-wise weighting $\balph$
is continuously valued in $\left[0 ; 1\right]^{H \times W}$ performing smooth
transitions between coding modes to avoid blocking artifacts.

CodecNet is similarly defined as a function $c$, which codes areas
$\nonstaticarea$ of $\code$ (selected through $\balph$) using information from
$\prediction$:
\begin{equation}
    R_{c},\ \hat{\mathbf{x}}_{t, c} = c\left(\alpha \odot \prediction, \alpha \odot \code\right).
\end{equation}
Element-wise matrix multiplication is denoted by $\odot$, $\hat{\mathbf{x}}_{t,
c} \in \mathbb{R}^{C \times H \times W}$ is the reconstruction of $\alpha \odot
\code$ and $R_c$ the associated rate. The same $\balph$ is used to multiply
all $C$ color channels. ModeNet is used to split $\code$ between what goes
through CodecNet and what is directly copied from $\prediction$, without
transmission. Thus the complete system output is:

\begin{equation}
    \sysoutput = (1 - \balph) \odot \prediction + c(\balph \odot \prediction, \balph \odot \code).
\end{equation}
This equation highlights that the role of $\balph$ is to zero areas from $\code$
before transmission to spare their associated rate. The whole system is trained
in an end-to-end fashion to minimize the rate-distortion trade-off:

\begin{equation}
    \mathcal{L}(\lambda) = \mathrm{D}(\sysoutput, \code) + \lambda (R_m + R_c),
    \label{eq:globalLoss}
\end{equation}

where $\mathrm{D}$ denotes a distortion metric. Following the CLIC20 P-frame
test conditions, the Multi Scale
Structural Similarity Metric (MS-SSIM)\cite{Wang03multi-scalestructural} is used:

\begin{equation*}
\mathrm{D}(\sysoutput, \code) = 1 - \text{MS-SSIM}(\sysoutput, \code).
\end{equation*}

As this work focuses on mode selection, a naive prediction $\prediction = \refframe = \prev$
is used. This allows to not add the burden of motion estimation to the system.
Results shown in this paper would still hold when working with a more relevant
prediction issued from motion compensation.

\newcommand{\expath}{\imagepath/2020_03_25--10_54_30/CoverSong_720P-3261_00185_detail/}
\newcommand{\exsize}{0.95}
\begin{figure*}[htb]
    \begin{subfigure}[b]{\exsize\columnwidth}
        \centering
        \includegraphics[width=\columnwidth]{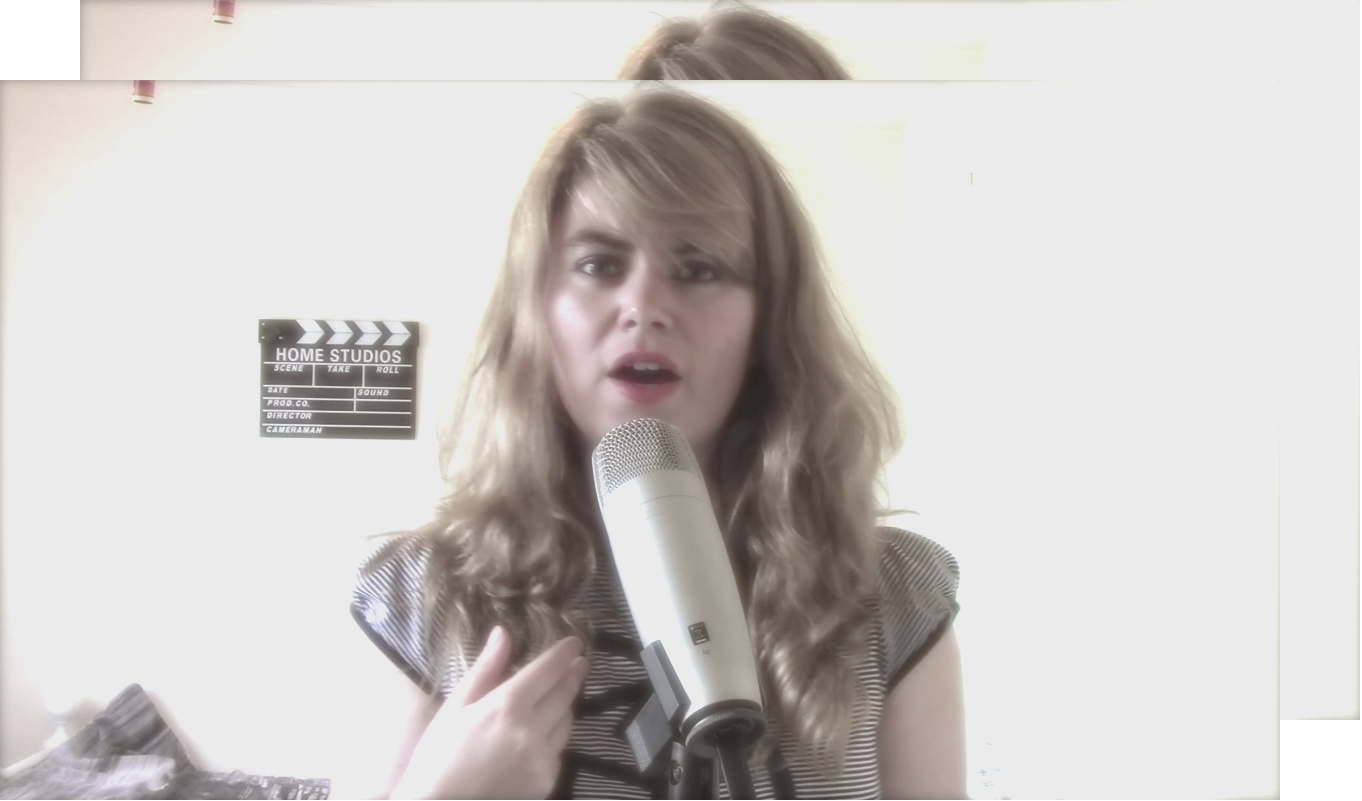}
        \caption{The pair of frames $(\prev,\code)$.}
        \label{ex:InputPairs}
    \end{subfigure}
    \hfill
    \begin{subfigure}[b]{\exsize\columnwidth}
        \centering
        \includegraphics[width=\columnwidth]{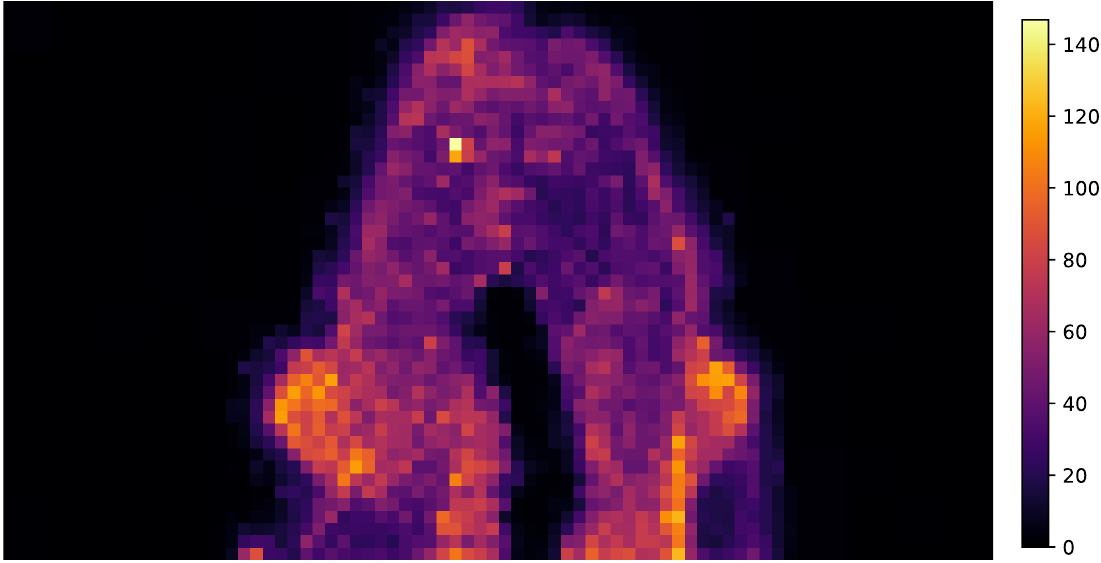}
        \caption{Spatial distribution of CodecNet rate in bits.}
        \label{ex:CodecNetRate}
    \end{subfigure}

    \begin{subfigure}[b]{\exsize\columnwidth}
        \centering
        \includegraphics[width=\columnwidth]{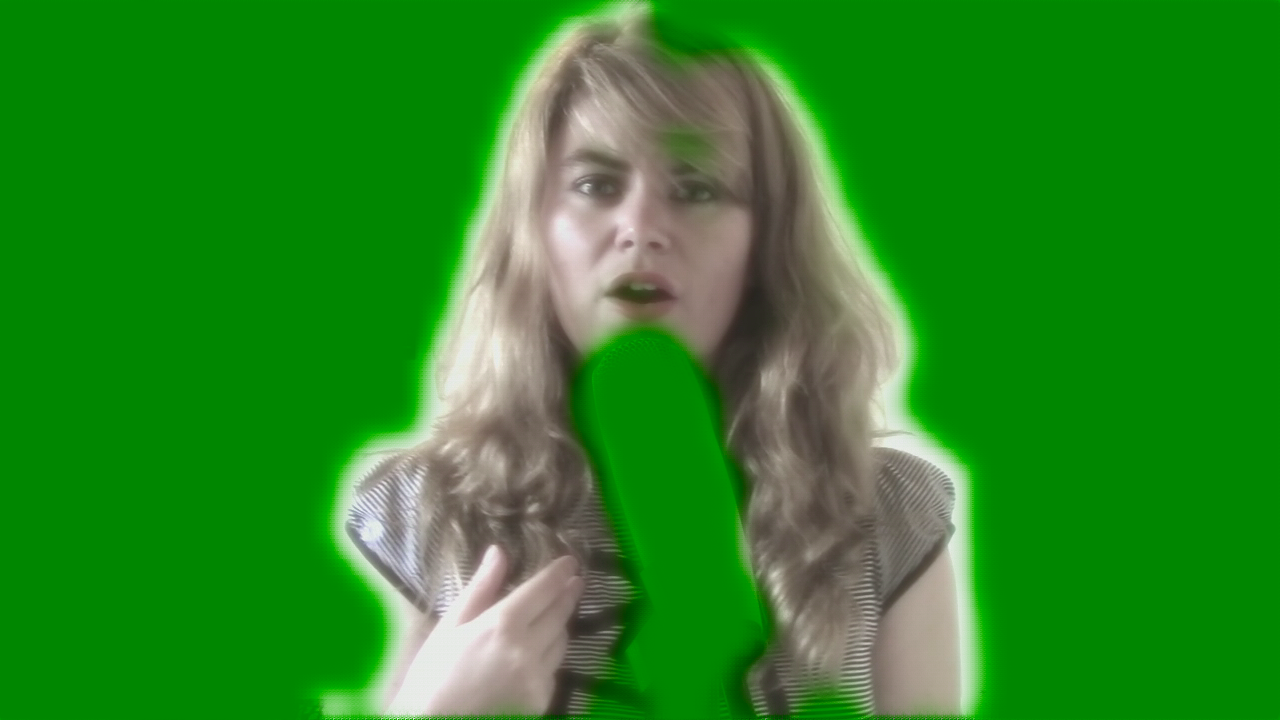}
        \caption{Areas selected for the CodecNet $\balph \odot \code$.}
        \label{ex:CodecNetPart}
    \end{subfigure}
    \hfill
    \begin{subfigure}[b]{\exsize\columnwidth}
        \centering
        \includegraphics[width=\columnwidth]{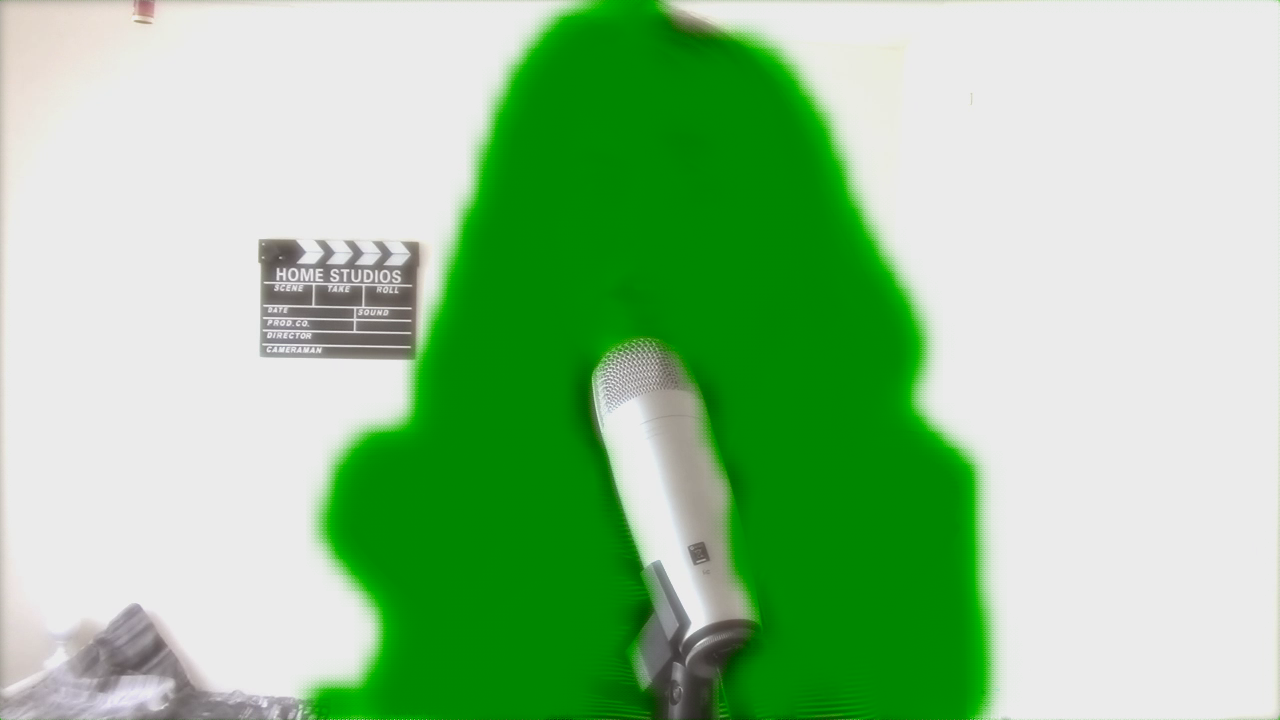}
        \caption{Areas selected for prediction copy $(1 - \balph) \odot \prediction$.}
        \label{ex:CopyPart}
    \end{subfigure}
    \caption{Details on the subdivision performed by ModeNet. The pair of frames
    $(\prev,\code)$ represents a singer moving in front of a static background.
    The microphone in the foreground is also motionless. Frame $\refframe = \prev$ is used
    as prediction $\prediction$.}
    \label{fig:Example}
\end{figure*}

\definecolor{imagesystems}{rgb}{0.161, 0.2, 0.361}
\definecolor{diffsystems}{rgb}{0.953, 0.654, 0.071}
\definecolor{conditionalsystems}{rgb}{0.859, 0.169, 0.224}
\definecolor{darkspringgreen}{rgb}{0.09, 0.45, 0.27}

\newcommand{\markscale}{1}
\newcommand{\markstarscale}{1.3}
\newcommand{\hevcmark}{star}
\newcommand{\hevcthickness}{thick}
\newcommand{\normalsystemmark}{square*}
\newcommand{\normalthickness}{thick}
\newcommand{\copylinestyle}{solid}
\newcommand{\copymark}{*}
\newcommand{\copythickness}{thick}
\newcommand{\normalsystemlinestyle}{densely dashed}
\newcommand{\hevclinestyle}{loosely dotted}

\newcommand{\resgraphwidth}{9cm}
\newcommand{\resgraphheigth}{7.25cm}
\newcommand{\resgraphxmin}{0}
\newcommand{\resgraphxmax}{0.2}
\newcommand{\resgraphymin}{12}
\newcommand{\resgraphymax}{25}

\definecolor{battleshipgrey}{rgb}{0.52, 0.52, 0.51}
\definecolor{davysgrey}{rgb}{0.33, 0.33, 0.33}
\definecolor{ashgrey}{rgb}{0.7, 0.75, 0.71}
\pgfplotsset{minor grid style={solid,battleshipgrey, dotted}}
\pgfplotsset{major grid style={solid, thick}}

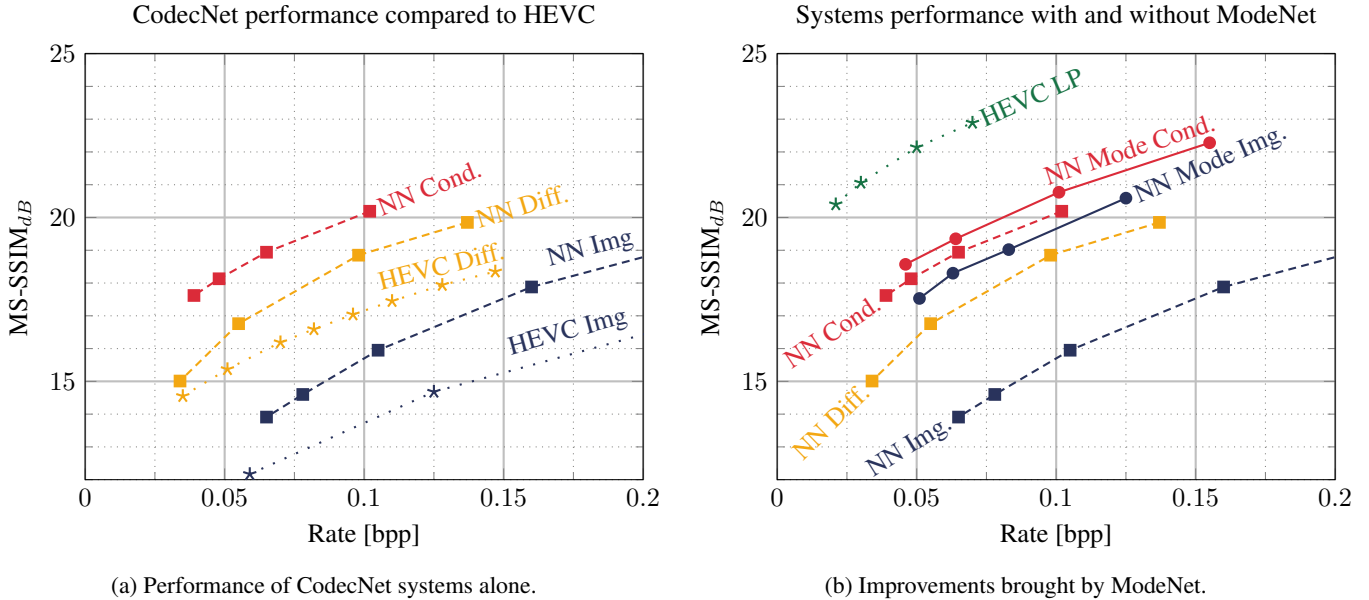
\begin{figure*}[htb]
    \begin{subfigure}[b]{1\columnwidth}
        \centering
        \begin{tikzpicture}
            \begin{axis}[
                grid= both ,
                xlabel = {Rate [bpp]} ,
                ylabel = {$\text{MS-SSIM}_{dB}$} ,
                xmin = \resgraphxmin, xmax = \resgraphxmax,
                ymin = \resgraphymin, ymax = \resgraphymax,
                ylabel near ticks,
                xlabel near ticks,
                width=\resgraphwidth,
                height=\resgraphheigth,
                x tick label style={
                    /pgf/number format/.cd,
                    fixed,
                    precision=2
                    },
                xtick distance={0.05},
                ytick distance={5},
                minor y tick num=4,
                minor x tick num=1,
                title=CodecNet performance compared to HEVC
            ]
                \addplot[\hevcthickness, \hevclinestyle, imagesystems, mark=\hevcmark, mark options={solid, scale=\markstarscale}] coordinates{
                    (0.059,12.17)
                    (0.125,14.68)
                    (0.240,17.36)
                    (0.426,20.34)
                }node [pos=0.45, sloped, anchor=south] {HEVC Img};

                \addplot[\normalthickness, \normalsystemlinestyle, imagesystems, mark=\normalsystemmark, mark options={solid, scale=\markscale}] coordinates {
                    (0.218,19.20)
                    (0.160,17.88)
                    (0.105,15.95)
                    (0.078,14.60)
                    (0.065,13.91)
                }node [pos=0.15, sloped, anchor=south] {NN Img};
                
                \addplot[\hevcthickness, \hevclinestyle, diffsystems, mark=\hevcmark, mark options={solid, scale=\markstarscale}] coordinates {
                    (0.035,14.55)
                    (0.051,15.37)
                    (0.070,16.19)
                    (0.082,16.59)
                    (0.096,17.04)
                    (0.110,17.45)
                    (0.128,17.94)
                    (0.147,18.35)
                }node [pos=0.9, sloped, anchor=south] {HEVC Diff.};

                \addplot[\normalthickness, \normalsystemlinestyle, diffsystems, mark=\normalsystemmark, mark options={solid, scale=\markscale}] coordinates {
                    (0.137,19.85)
                    (0.098,18.85)
                    (0.055,16.76)
                    (0.034,15.01)
                }node [pos=0, sloped, anchor=west] {NN Diff.};

                \addplot[\normalthickness, \normalsystemlinestyle, conditionalsystems, mark=\normalsystemmark, mark options={solid, scale=\markscale}] coordinates {
                    (0.102,20.19)
                    (0.065,18.94)
                    (0.048,18.13)
                    (0.039,17.62) 
                }node [pos=0, sloped, anchor=west] {NN Cond.}; 
            \end{axis}
        \end{tikzpicture}
        \caption{Performance of CodecNet systems alone.}
        \label{AnchorResults}
    \end{subfigure}
    \hfill
    \begin{subfigure}[b]{1\columnwidth}
        \centering
        \begin{tikzpicture}
            \begin{axis}[
                grid= both ,
                xlabel = {Rate [bpp]} ,
                ylabel = {$\text{MS-SSIM}_{dB}$} ,
                xmin = \resgraphxmin, xmax = \resgraphxmax,
                ymin = \resgraphymin, ymax = \resgraphymax,
                width=\resgraphwidth,
                height=\resgraphheigth,
                ylabel near ticks,
                xlabel near ticks,                
                x tick label style={
                    /pgf/number format/.cd,
                    fixed,
                    precision=2
                    },
                xtick distance={0.05},
                ytick distance={5},
                minor y tick num=4,
                minor x tick num=1,
                title=Systems performance with and without ModeNet
            ]

                \addplot[\normalthickness, \normalsystemlinestyle, imagesystems, mark=\normalsystemmark, mark options={solid, scale=\markscale}] coordinates {
                    (0.218,19.20)
                    (0.160,17.88)
                    (0.105,15.95)
                    (0.078,14.60)
                    (0.065,13.91)
                }node [pos=1, sloped, anchor=east] {NN Img.};
                    
                \addplot[\normalthickness, \normalsystemlinestyle, diffsystems, mark=\normalsystemmark, mark options={solid, scale=\markscale}] coordinates {
                    (0.137,19.85)
                    (0.098,18.85)
                    (0.055,16.76)
                    (0.034,15.01)
                }node [pos=1, sloped, anchor=east] {NN Diff.};

                \addplot[\normalthickness, \normalsystemlinestyle, conditionalsystems, mark=\normalsystemmark, mark options={solid, scale=\markscale}] coordinates {
                    (0.102,20.19)
                    (0.065,18.94)
                    (0.048,18.13)
                    (0.039,17.62) 
                }node [pos=1, sloped, anchor=east] {NN Cond.};
                
                \addplot[\copythickness, \copylinestyle, imagesystems, mark=\copymark, mark options={solid, scale=\markscale}] coordinates {
                    (0.125,20.59)
                    (0.083,19.02)
                    (0.063,18.30)
                    (0.051,17.53)
                }node [pos=0, sloped, anchor=west, yshift=-0.05cm] {NN Mode Img.};                          


                \addplot[\copythickness, \copylinestyle, conditionalsystems, mark=\copymark, mark options={solid, scale=\markscale}] coordinates {
                    (0.155,22.28)
                    (0.101,20.77)
                    (0.064,19.35)
                    (0.046,18.57)
                }node [pos=0.2, sloped, anchor=south] {NN Mode Cond.};

                \addplot[\hevcthickness, \hevclinestyle, darkspringgreen, mark=\hevcmark, mark options={solid, scale=\markstarscale}] coordinates {
                    (0.021,20.40)
                    (0.030,21.06)
                    (0.050,22.14)
                    (0.070,22.89)
                }node [pos=1, sloped, anchor=west] {HEVC LP};        
            \end{axis}
        \end{tikzpicture}
        \caption{Improvements brought by ModeNet.}
        \label{ModeNetResults}
    \end{subfigure}
    \caption{Rate-distortion performance of the systems. All systems are
    evaluated on CLIC20 P-frame validation dataset. Quality metric is
    $\text{MS-SSIM}_{dB} = -10 \log_{10} (1 - \text{MS-SSIM})$ (the higher the
    better). Rate is indicated in bits per pixel (bpp). Img. denotes image,
    Diff. difference, Cond. conditional and HEVC LP is HEVC in low-delay P
    configuration.}
    \label{fig:results}
\end{figure*}

\subsection{Networks architecture}

\label{subsec:detailArchitecture}
Both ModeNet and CodecNet networks are built from standard AE-HP blocks
described in Fig. \ref{RawBlocksDiagrams}. The ModeNet role is to process the
previous and current frames to transmit the pixel-wise weighting $\balph$. It is
implemented as a lightweight AE-HP system (\textit{cf.} Fig.
\ref{ModeNetDiagram}), with $\prev$ and $\code$ as inputs. A bias of $0.5$ is
added to the output as it makes training easier. To assure that $\balph \in
\left[0, 1\right]^{H \times W}$ a clipping function is used.There are 200~000
parameters in ModeNet, which represents around 10~\% of CodecNet number of
parameters.

In order to transmit pixels in $\nonstaticarea$, three different configurations
of CodecNet are investigated. Two of them are based on the architecture depicted
in Fig. \ref{CodecNetImageDiffDiagram}. They consist in either plain image
coding of $\code$ or in difference coding of ($\code - \prediction$) (prediction
error coding). Last configuration is conditional coding denoted as $(\code \mid
\prediction)$. shown Fig. \ref{CodecNetConditionalDiagram}. This configuration
theoretically results in better performance. Indeed, from a source coding
perspective:

\begin{equation}
    \entropy(\code \mid \prediction) \leq \min\left(\entropy(\code),\ \entropy(\code - \prediction)\right).
    \label{eq:entropy}
\end{equation}

Therefore, coding $\code$ while retrieving all information from $\prediction$
results in less information to transmit than difference or image coding. 

\section{Network training}
All networks are trained in an end-to-end fashion to minimize the global loss
function stated in eq. \eqref{eq:globalLoss}. Non-differentiable parts are
approximated as in Ball\'e's work
\cite{DBLP:conf/iclr/BalleLS17,DBLP:conf/iclr/BalleMSHJ18} to make training
possible. End-to-end training allows ModeNet to learn to partition $\code$,
without the need of an auxiliary loss or a hand-crafted criterion. Due to the
competition between signal paths, some care is taken when training. The training
process is composed of two stages:

\textbf{Warm-up.} Training of CodecNet only (\textit{i.e.} ModeNet weights are
frozen). Unlike copy, CodecNet is not immediately ready to process its input.
Thus CodecNet has to be trained for a few epochs so the competition between copy
and CodecNet is relevant.

\textbf{Alternate training.} Alternate training of ModeNet and CodecNet, one
epoch for each (\textit{i.e.} the other network weights are frozen).

The training set is constructed from the CLIC20 P-frame training dataset
\cite{CLIC20_web_page}. Half a million $256 \times 256$ pairs of crops are
randomly extracted from consecutive frames. The batch size is set to 8 and an
initial learning rate of $10^{-4}$ is used. The learning rate is divided by 5 at
50~\% and 75~\% of the training.

\section{Mode Visualisation}

This sections details the processing of a pair of frames $(\prev,\code)$ by the
proposed system. Frames are from the sequence $\textit{CoverSong\_720P-3261}$
extracted from the CLIC20 P-frame dataset. The system used for generating the
visuals is implemented as in Fig. \ref{CompleteSystemDiagrams}.

Figure \ref{ex:InputPairs} shows the inputs of ModeNet. They are encoded and
decoded as the pixel-wise weighting $\balph$. The value of $\balph$ tends to be
zero\footnote{As images are in YUV format, all-zero areas appear in green}
for pixels in $\staticarea$ \textit{i.e.} when copying $\prediction$ results in
a better rate-distortion cost than transmission through CodecNet. $\staticarea$
corresponds to static areas in $(\prev, \code)$ as the background and the
microphone, which are well captured by $\balph$. This areas are shown in Fig.
\ref{ex:CopyPart}.

CodecNet selected inputs are $\balph \odot \code$ and $\balph \odot \prediction$
depicted in Fig. \ref{ex:CodecNetPart}. Copying areas of the prediction
$\prediction$ allows to zero areas in $\code$ which prevents CodecNet to spend
rate for these areas. Figure \eqref{ex:CodecNetRate} shows the spatial
distribution of the rate in CodecNet and clearly highlights this behavior.

In this example, the rate associated to $\balph$ is 0.005 bit per pixel (bpp).
This shows that ModeNet is able to convey a smooth partitioning of an arbitrary
number of objects for a marginal amount of rate.

\section{Experimental Results}

Performance improvements brought by ModeNet are assessed on the CLIC20 P-frame
validation set, under the challenge test conditions. In order to obtain
RD-curves, each system is learnt with different $\lambda$. Results are gathered
in Fig.~\ref{fig:results}. For the sake of brevity, systems denoted as
\textit{NN Mode X} are complete systems (\textit{cf.} Fig.
\ref{CompleteSystemDiagrams}) composed of both ModeNet and CodecNet in coding
configuration X. Similarly, systems \textit{NN X} denotes CodecNet only system
without ModeNet (\textit{i.e.} no copy possibility: $\balph$ is an all-ones matrix).

\subsection{Anchors}

CodecNet performance is assessed by training and evaluating it without ModeNet,
meaning that $\code$ is completely coded through CodecNet. The three
configurations of CodecNet (\textit{cf.} section \ref{subsec:detailArchitecture}
and Fig. \ref{fig:four figures}) are tested. The image configuration is
compared with HEVC in All Intra configuration. Difference configuration is compared
with HEVC coding the pre-computed difference image. For both comparison, HEVC
encodings are performed with the HM 16.20 reference software. Results in terms of
MS-SSIM versus the rate are shown in Fig. \ref{AnchorResults}. CodecNet achieves
consistently better performance than HEVC for both configurations across all
bitrates, proving its competitiveness.

Conditional coding achieves better performance than both difference and image
coding as expected from eq. \eqref{eq:entropy}. This shows the relevance of
performing conditional coding relative to difference coding.

\subsection{Performances of ModeNet-based systems}

Performances of ModeNet-based systems are shown Fig. \ref{ModeNetResults}.
Using ModeNet increases the performance of both image and conditional coding.
Image coding of $\code$ alone does not have any information about the previous
frame. Thus, adding ModeNet and the possibility of copying areas of
$\prediction$ results in an important increase of the performance. 

Interestingly, NN Mode Image achieves significantly better results that NN
Difference. As illustrated in Fig. \ref{fig:Example}, $\staticarea$ tends to
represent the areas similar in $(\prediction, \code)$, which are well
handled by difference coding. Thus, performance gap between NN Mode
Image and NN Difference arises on $\nonstaticarea$, where image coding
outperforms difference coding. 

An ideal conditional coder is able to retrieve all informations about
$\code$ in $\prediction$ making $\prediction$ copy useless. However, leveraging
all information in $\prediction$ is not possible for a neural network with
reduced complexity. There are still areas for which $\prediction$ copy provides
a smaller rate-distortion cost than transmission. Thus using ModeNet to identify
them improves performance.

To better appreciate the results, HEVC low-delay P (LP) performance is presented.
HEVC LP codes $\code$ with $\prev$ as reference frame and is able to perform motion
compensation to obtain a relevant prediction. Consequently, it outperforms all
other systems which are constrained to directly use $\prev$ as their prediction, without motion compensation.

Using ModeNet with the best CodecNet configuration (conditional coding) allows
to decrease the rate by 40~\% compared to difference coding for the whole frame.
Even though this gap would decrease when working with a motion compensated
prediction, we believe that using ModeNet to arbitrate between conditional
coding of $(\code \mid \prediction)$ and copy of $\prediction$ would improve
most learned video coding methods, which still uses difference coding for the
whole frame.

\section{Conclusion and Future Works}

In this paper, we propose a mode selection network which learns to transmit a
partitioning of a frame to code, allowing to choose among different coding
methods pixel-wise. ModeNet benefits are illustrated under a P-frame coding
task. It is shown that coding the prediction error is not necessarily the best
choice and using ModeNet to select better coding methods significantly
increase performance.

This paper shows that the proposed ModeNet performs a smooth partitioning of an
arbitrary number of areas in a frame, for a marginal rate and complexity
overhead. It can be generalized to other coding schemes to leverage competition of
complementary coding modes, which is known to be one of the most powerful tools
in classical video coding.

An extension of this work is to use motion information to improve the
prediction process. As the proposed method outperforms residual coding, having a
competitive motion compensated prediction would result in compelling
performance.

\bibliographystyle{IEEEbib}
\bibliography{refs}

\end{document}